\documentclass[]{interact}

\usepackage{epstopdf}
\usepackage[caption=false]{subfig}

\usepackage{natbib}
\bibpunct[, ]{(}{)}{;}{a}{}{,}
\renewcommand\bibfont{\fontsize{10}{12}\selectfont}

\theoremstyle{plain}

\theoremstyle{definition}

\theoremstyle{remark}

\usepackage[dvipsnames]{xcolor}
\newcommand{\revision}[1][]{#1}

\usepackage{standalone}
\usepackage{tikz}

\usepackage{algorithm}
\usepackage{program}
\usepackage{amsmath}
\usepackage{multirow}
\usepackage{booktabs}

\usepackage{placeins}

\newcommand{\bftab}{\fontseries{b}\selectfont}

\newcommand{\anoncmd}[2]{%
\ifdefined\anonymise%
#2%
\else%
#1%
\fi%
}

\newcommand{\figref}[1]{\figurename~\ref{#1}}

\usepackage[hidelinks]{hyperref}

\begin{document}

\title{Rapid and robust endoscopic content area estimation: A lean GPU-based pipeline and curated benchmark dataset}


\anoncmd{
\author{
\name{Charlie Budd\textsuperscript{a}\thanks{CONTACT Charlie Budd. Email: charles.budd@kcl.ac.uk}, Luis C. Garcia-Peraza-Herrera\textsuperscript{a}, Martin Huber\textsuperscript{a}, Sebastien Ourselin\textsuperscript{a,b}, and Tom Vercauteren\textsuperscript{a,b}}
\affil{
\textsuperscript{a}King's College London, UK\\
\textsuperscript{b}Hypervision Surgical Ltd, UK}
}
}{\author{Anonymous}}
\maketitle

\begin{abstract}
Endoscopic content area refers to the informative area enclosed by the dark, non-informative, border regions present in most endoscopic footage.
The estimation of the content area is a common task in endoscopic image processing and computer vision pipelines. 
Despite the apparent simplicity of the problem, several factors make reliable real-time estimation surprisingly challenging.
The lack of rigorous investigation into the topic combined with the lack of a common benchmark dataset for this task has been a long-lasting issue in the field.
In this paper, we propose
two variants of a lean GPU-based computational pipeline combining edge detection and circle fitting.
The two variants differ by relying on handcrafted features, and learned features respectively to extract content area edge point candidates.
We also present a first-of-its-kind dataset of manually annotated and pseudo-labelled content areas across a range of surgical indications.
To encourage further developments, the curated dataset, and an implementation of both algorithms, has been made public (\anoncmd{\url{https://doi.org/10.7303/syn32148000}, \url{https://github.com/charliebudd/torch-content-area}}{anonymized url}).
We compare our proposed algorithm with a state-of-the-art U-Net-based approach and demonstrate significant improvement in terms of both accuracy (Hausdorff distance: 6.3 px versus 118.1 px) and computational time (Average runtime per frame: 0.13 ms versus 11.2 ms).
\end{abstract}


\section{Introduction}
\subsection{Endoscopic content area}
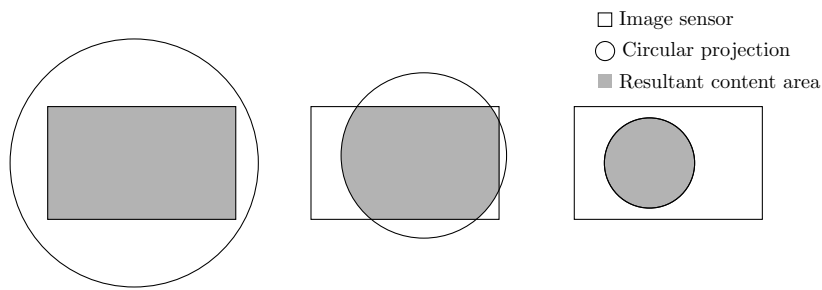
\begin{figure}[tbh]
\centering
\begin{tikzpicture}[scale=0.5,  every node/.style={scale=0.7}]  

\useasboundingbox (-2,-2) rectangle (21,6);

\begin{scope}
    \clip (0,0) rectangle (5,3);
    \fill[
        fill=gray!60
    ] (2.3,1.5) circle (3.3);
\end{scope}
\draw (0,0) rectangle (5,3);
\draw (2.3,1.5) circle (3.3);

\begin{scope}
    \clip (0+7,0) rectangle (5+7,3);
    \fill[
        fill=gray!60,
    ] (3+7,1.7) circle (2.2);
\end{scope}
\draw (0+7,0) rectangle (5+7,3);
\draw (3+7,1.7) circle (2.2);

\begin{scope}
    \clip (0+14,0) rectangle (5+14,3);
    \fill[
        fill=gray!60,
        draw=black
    ] (2+14,1.5) circle (1.2);
\end{scope}
\draw (0+14,0) rectangle (5+14,3);
\draw (2+14,1.5) circle (1.2);

\matrix [below left] at (current bounding box.north east) {
  \node [style={shape=rectangle, draw=black},label=right:Image sensor] {}; \\
  \node [style={shape=circle, draw=black},label=right:Circular projection] {}; \\
  \node [style={shape=rectangle, fill=gray!60},label=right:Resultant content area] {}; \\
};

\end{tikzpicture}
\caption{A diagram showing examples of the formation of the content area as the intersection of the circular image projection and the rectangular image sensor. The leftmost content area forms a rectangle, while the rightmost content area forms a complete circle. The central example, however, forms a more complex shape formed from straight lines and circular arcs.} 
\label{fig:intersection_diagram}
\end{figure}

\begin{figure}[tbh]
\centering
\resizebox*{10cm}{!}{\includegraphics{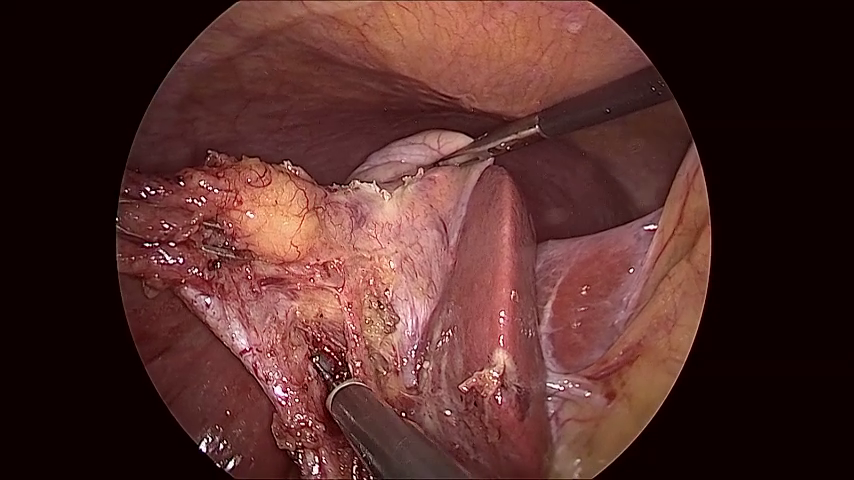}}
\caption{A frame from a laparoscopic procedure taken from the Cholec80 dataset \citep{twinanda16}. The image shows a clear example of a circular endoscopic content area. The content area is bright and well centred, the border region is dark and almost noise free, and the edge between them is clear and sharp. The content area is incomplete in that one or more sections of the content area extend beyond the extents of the frame. Examples such as this, where the top and bottom of the content area are cropped, are very common.} 
\label{fig:easy_example}
\end{figure}

In normal commercial cameras, the optics create a circular projection which fully covers the image sensor, resulting in a full rectangular image.
In minimally invasive intervention, however, an endoscope is often used to allow an external camera to view inside the patient. 
The size restriction of the endoscope \revision[constrains] the field of view that can be captured by the optics.
This reduction in the potentially visible area, and the critical nature of the application, make it desirable for the surgeon to see a large portion of the optical field of view. 
As illustrated in \figref{fig:intersection_diagram} and \figref{fig:easy_example}, a typical trade-off is thus to allow the circular projection to fall so that part of the image sensor lies outside the projection.
This maximises what is visible to the surgeon, but results in dark, non-informative regions near the edge of the image.
We define the endoscopic content area as the informative region of the image, which is formed by the intersection of the circular image projection with the image sensor.
We further define the border as being the (non-informative) sensor area not covered by the circular image projection.

\subsection{Content area estimation}
Estimation of the content area is a common task in endoscopic image processing and computer vision.
We define the task as the estimation of the geometric shape of the content area, not just a pixel wise segmentation.

\revision[The estimation forms a foundational building block when constructing a geometric understanding of an endoscopic image. For example, determining where the endoscopic view is centred, or if a detected tool is obscured by organic tissue or is simply exiting the content area. Both of these tasks are vital when considering robotic control, such as autonomous control of the endoscope \citep{gruijthuijsen2022}.
The knowledge of the content area may also be useful when training and using deep learning vision models. The regions outside of the content area contain non-informative details in the form of noise and text overlays. Indeed, these details may, in fact, serve to bias a model. For example, the information outside of the content area may be characteristic of the endoscope used during the intervention. A model trained to detect the type of intervention may learn to detect the characteristic non-content area information present in the training data. Masking out these details could simplify the input and remove such sources of bias. 
More speculatively, when training a segmentation or object detection model, the loss function could be modified so it does not penalise predictions made outside of the content area. In this way, the task to be learned may be simplified as the model need not learn to correctly classify these regions. At inference time, any activations in these regions may be discarded.
Additionally, as the content area only takes up a portion of the image, the amount of required computation can be reduced by skipping border regions when performing inference for time-critical applications.]

For any follow-up task to be able to rely on the estimated content area, a high level of robustness must be achieved under all expected conditions.
While still important, precision is less of a concern, as a content area found to be slightly off from the true content area will likely have little consequence on subsequent processing.
To utilise content area estimation in a real-time setting, the estimation would ideally use minimal computing resources and processing time.
Thereby leaving these resources available for follow-up tasks.

As image sensor technology improves, and the manufacture of sensitive compact image sensors becomes cheaper, it may become increasingly common to mount the sensor on the end of the endoscope, known as chip-on-tip. 
This may remove the circular border artefact currently mostly prominent when using proximal cameras.
Should chip-on-tip endoscopes become the only norm, estimation of the content area may become less critical.
Until such a time, it will remain important to be able to efficiently and reliably detect the border.
Several chip-on-tip endoscope manufacturer, especially in the flexible endoscopy field, also continue to opt for an endoscope design with incomplete content area. 
Finally, the ability to exploit historical endoscopic imaging data also warrants the availability of robust content area estimation algorithms.

\subsection{Challenges in content area estimation}
\begin{figure}[tb!]
\centering
\subfloat[A saturated area at the edge of the content area bleeds into the border.]{%
\resizebox*{7cm}{!}{\includegraphics{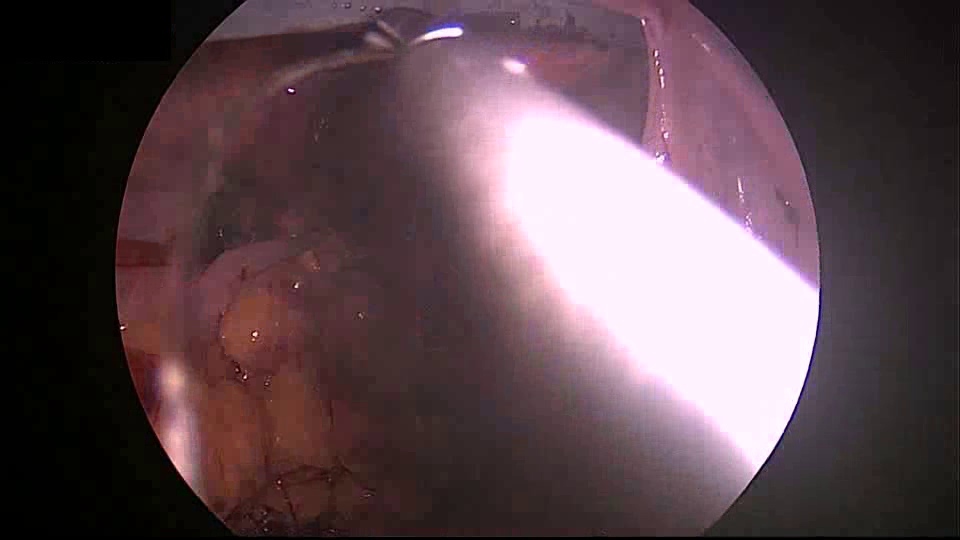}}}%
\hspace{5pt}
\subfloat[A dark content area leaves only a small segment of the circle visible.]{%
\resizebox*{7cm}{!}{\includegraphics{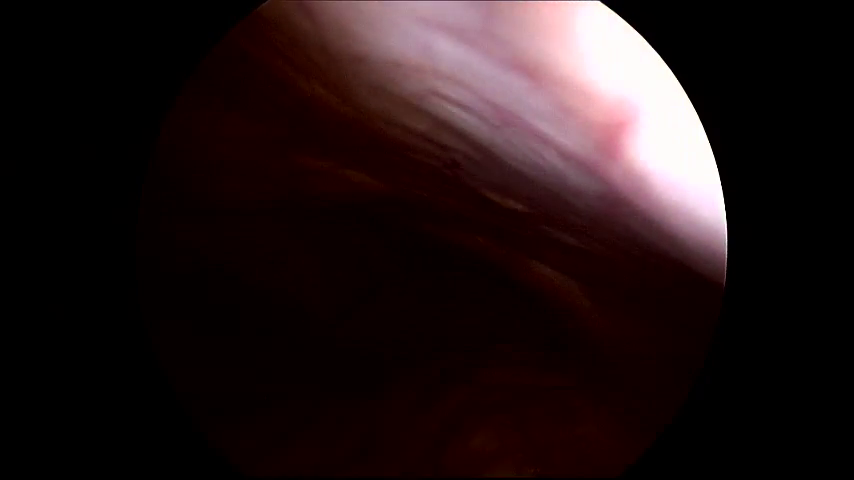}}} \\
\vspace{-1ex}
\subfloat[A partially cropped circle combined with a black overlay.]{%
\resizebox*{7cm}{!}{\includegraphics{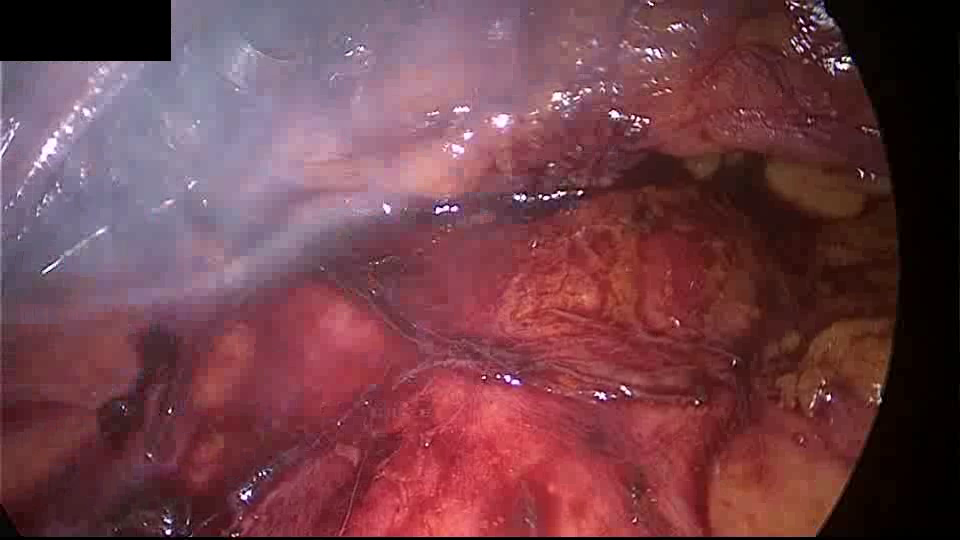}}}
\hspace{5pt}
\subfloat[Text overlays the content area and border.]{%
\resizebox*{7cm}{!}{\includegraphics{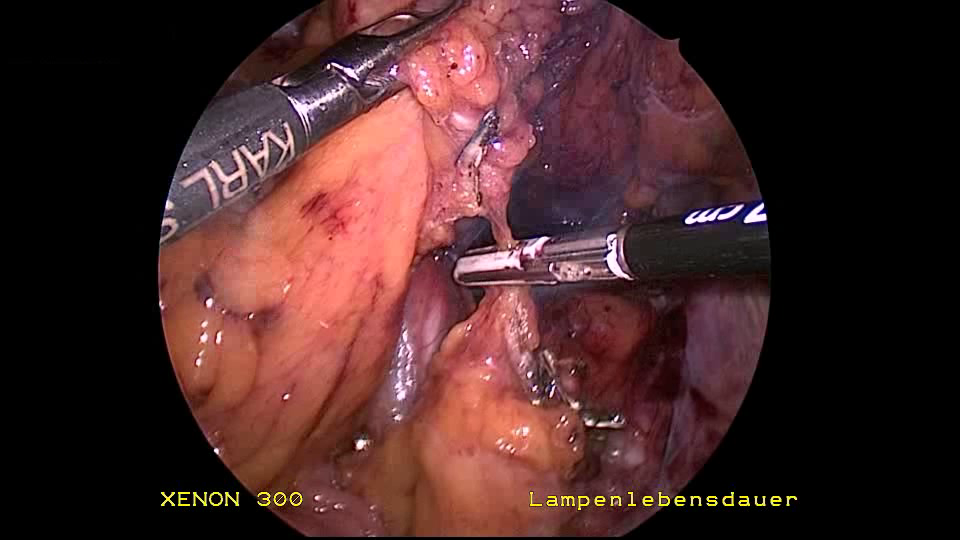}}} \\
\vspace{-1ex}
\subfloat[Structure within the top right quadrant of the content area appears as a misleading circle segment.]{%
\resizebox*{7cm}{!}{\includegraphics{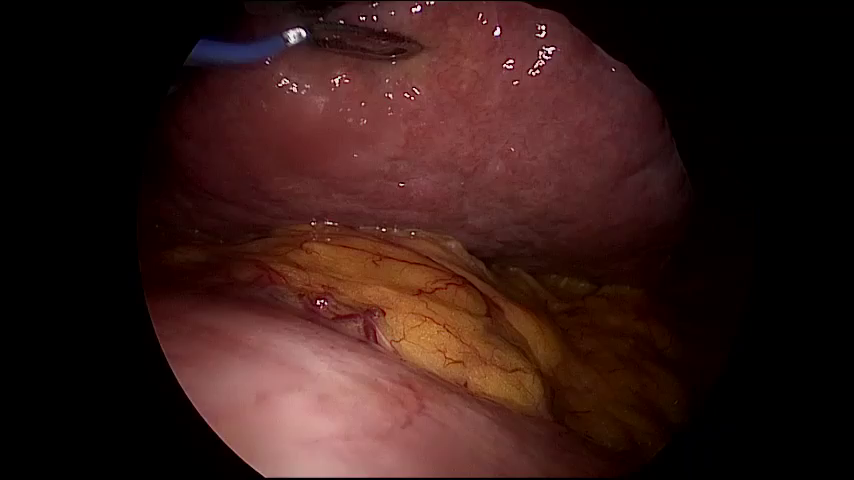}}}
\hspace{5pt}
\subfloat[A dark content area combined with a mostly cropped border provides a truly challenging example.]{%
\resizebox*{7cm}{!}{\includegraphics{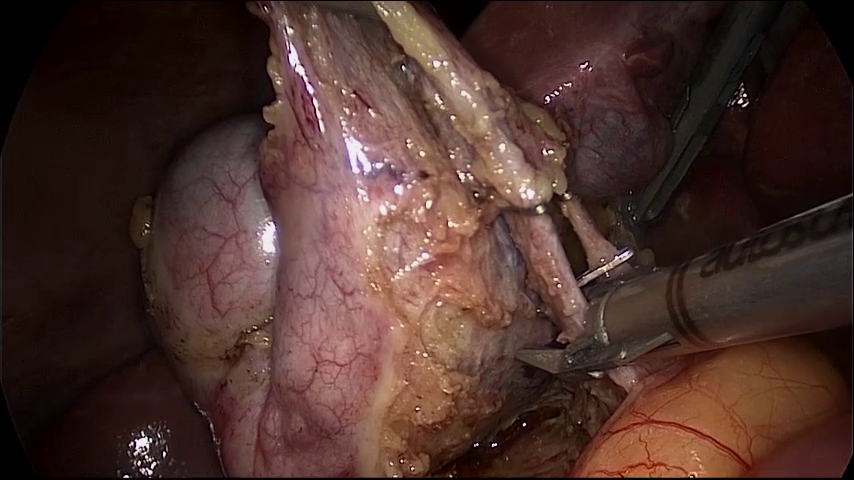}}}
\caption{A selection of examples taken from our hand annotated dataset, chosen to portray some of the adverse features faced during endoscopic content area estimation.} 
\label{fig:hard_examples}
\end{figure}

Delineation between the border and the content area of the image is made non-trivial by a few factors.
\figref{fig:hard_examples} shows a selection of endoscopic images demonstrating some of these difficulties.
Firstly, while the border is generally a uniform black, a fair amount of low level noise is often observed, and imperfections in the scopes optics can result in aberrations such as bright spots, diffuse light bleeding outside of the content area, or imperfect circles. 
Secondly, the image within the content area may be adverse in that it can have low brightness or contain within it a secondary circular oculus, such as when the tip of the endoscope is only partially inserted through a trocar.
Thirdly, while the circular image projection is generally centred around the middle of the image, it can in fact be significantly offset from the centre and its radius can fall within in a large range, even passing beyond the horizontal extent of the image for much of the image height.
The spatial position and size of the circular image projection may also be surprisingly dynamic throughout an intervention, varying due to mechanical stresses placed through the endoscope and as the operator adjusts the zoom level on the camera. 
Finally, there can exist additional overlays such as secondary camera feeds, logos, and text.

\subsection{Related work}
Despite the fundamental nature of endoscopic content area estimation, there has been little published work dedicated to the topic.
One possible reason for this is that the problem may be assumed to be solved by trivial methods, such as applying an intensity threshold. 
While this is indeed the case for the majority of images, we find that real endoscopic video often contains difficult cases which are missed by such approaches.
Another potential explanation may relate to the absence of established datasets to develop and evaluate content area estimation algorithms.

Previous dedicated investigation into the problem of endoscopic content area estimation was performed by \cite{munzer13}. 
The contribution is provided without an open-source implementation or open-access dataset, thereby limiting the potential impact of their work.
Their algorithm checks for a black vertical strip on the left and right extremes of the image, from this it can be assumed that they worked with images which contained the full horizontal extent of the content area. 
In practice, we find this to not always be the case (see examples c, and f in \figref{fig:hard_examples}), and that these cases are often the most difficult to detect correctly.

Other publications exist that require content area estimation as a preprocessing step, however, as the estimation is not the focus of the publication, the solution and presentation lack depth and rigour.
For example, \cite{huang21} investigate the removal of the non-informative regions to improve CNN performance by cropping the polar representation of the image.
Despite being aware of the contribution by \cite{munzer13}, they choose to hand select clips with static content areas, which are then manually defined. 
Another example is provided by \cite{huber22}. Whilst investigating deep homography estimation in surgical scenes, they choose to develop their own content area estimation method.
The method is based on a \revision[U-Net \citep{ronneberge15}] trained to provide a pixel wise segmentation of the content area.
A Sobel filter is then applied to extract points along the edge of the segmentation.
A least squares fitting is then used to fit a circle to a random subset of these edge points.
A final example is provided by \cite{gruijthuijsen2022}. In their work on robotic endoscope control, they develop a content area estimation method which relies primarily on a HSV filter followed by some morphological processing. 
Both \cite{huber22}, and \cite{gruijthuijsen2022} provide implementations of their approaches \revision[on GitHub\footnote{\url{https://github.com/RViMLab/endoscopy} \citep{huber22}}$^,$\footnote{\url{https://github.com/luiscarlosgph/endoseg} \citep{gruijthuijsen2022}}], the results from which are included in the results section.

Looking more broadly, we find more recent publications on single dominant circle estimation for manufacturing applications. 
Whilst investigating bottle mouth opening estimation, \cite{zhou2022} suggest using radial scan lines centred at the barycentre of pixels above an intensity threshold, before using model fitting in polar coordinates to remove incorrect edge points.
\cite{splett2019} investigate high precision circle fitting for optical fibre cross-sections. The manufacturing tolerances they are considering call for a much higher degree of precision than is required in our application, and so this work is not incorporated here.
Both these applications have much clearer circles to detect and have more lax time and computational constraints, so their methods are less applicable to endoscopic content area estimation.

\subsection{Contributions}
In this work, we
provide a
rigorous investigation into content area estimation.
We propose two variants of a lean GPU accelerated computational pipeline
for performing content area estimation.
Our pipeline comprises feature extraction, content area edge candidate detection, and circle fitting. Our two variants differs in using handcrafted features or learned features for edge detection purposes.
Care is taken in both cases to minimise the computational overhead.
We also present the first open-access dataset for evaluating content area estimation methods.
We compare our algorithms with an openly available U-Net based content area detection method on the presented dataset and demonstrate significant improvements in terms of both accuracy, robustness, and runtime.
Both the dataset and implementations of our algorithms are made publicly available to encourage further work.

\section{Materials and methods}

\subsection{Dataset}
\begin{figure}[tb!]
\centering
\resizebox*{7cm}{!}{\includegraphics{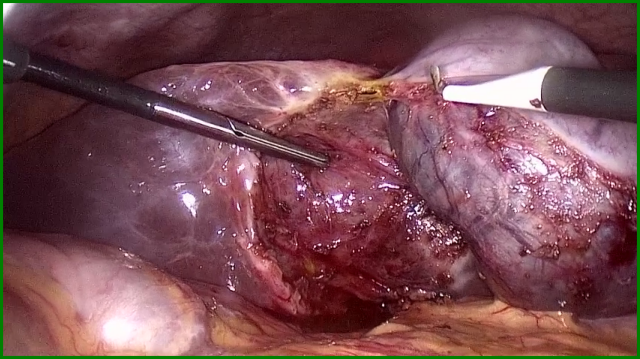}}
\hspace{-6pt}
\resizebox*{7cm}{!}{\includegraphics{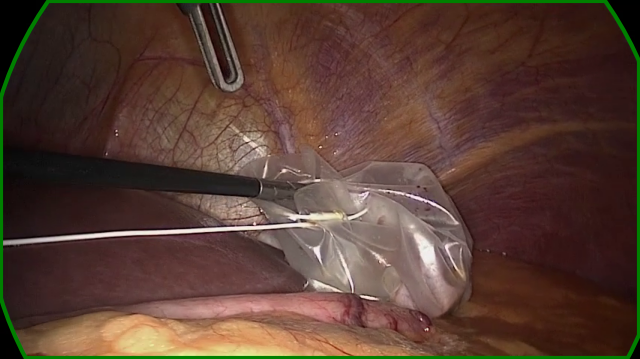}}
\resizebox*{7cm}{!}{\includegraphics{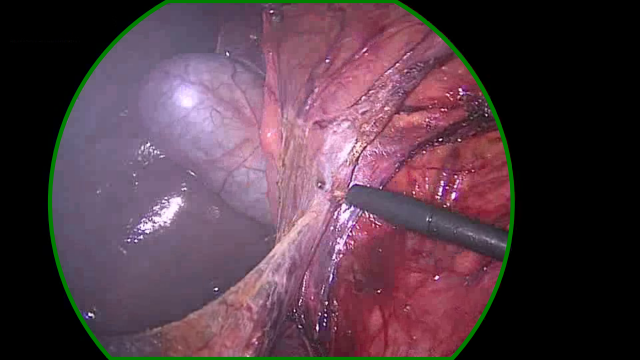}}
\hspace{-6pt}
\resizebox*{7cm}{!}{\includegraphics{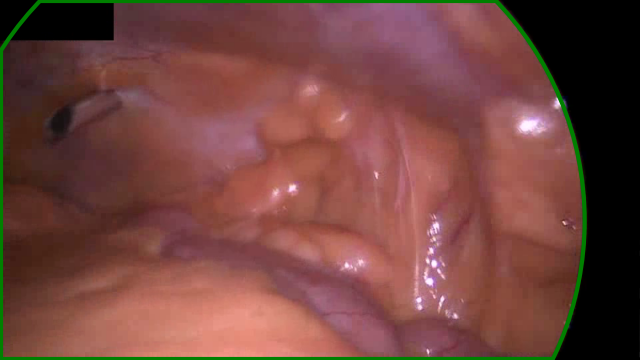}}
\caption{\revision[A selection of samples from the CholecECA and RobustECA datasets, showing the edges of the annotated content areas. From top-left in clockwise order: No border regions visible, content area takes up the full image rectangle; Border regions are visible in all four corners; An almost full circle is shown with around half the image area covered by a border region; The right-hand edge is covered with a border region, the upper-left-hand corner also show some border region, although it is only just visible due to the blacked out secondary video-stream.]} 
\label{fig:ground_truths}
\end{figure}

We created a manually annotated content area dataset based on Cholec80 \citep{twinanda16}, and RobustMIS \citep{tobias20}, two popular, publicly available, laparoscopy video datasets.
Combined, these two source datasets contain videos of 110 different laparoscopic procedures (80 from Cholec80, and 30 from RobustMIS).
We take the raw videos from each of these datasets and sample them to extract 3500 images, (2000 from Cholec80 and 1500 from RobustMIS). 
These images are visually inspected for border regions. 
If any are found, points are manually placed along the circular arcs of the edge between the content area and border regions.
A circle is then fitted to these edge points.
Each image which contains border regions are also cropped to within the content area to provide an additional sample, without border regions. 
This serves to better balance the numbers of images with and without border regions, as well as increasing the size of the dataset \revision[to 6923, (3929 from Cholec80 and 2994 from RobustMIS).] 
The origin dataset, video number, and frame number of each sample is recorded to facilitate the testing of future methods which make use of temporal information. These two datasets are dubbed CholecECA and RobustECA respectively. \revision[Some examples of the ground truth annotations may be seen in \figref{fig:ground_truths}.]

In addition to the hand annotated dataset, we created a much larger pseudo-labelled dataset, dubbed PseudoECA, for training the learned edge scoring method presented in this paper. The data is sampled from Cholec80 at a frame every 2 seconds, producing 92,310 samples. The content area of each image is inferred using the handcrafted edge scoring variant of the algorithm presented in this paper. Table~\ref{dataset-table} shows a list of the curated datasets.

\begin{table}
\tbl{A table showing the size, source, and annotation method of our curated datasets}
{
    \begin{tabular}{lcccc} 
    \toprule
    Name            & Source        & Annotations       & No. Samples   \\
    \midrule
    CholecECA       & Cholec80      & Manual            & \revision[3929] \\
    RobustECA       & RobustMIS     & Manual            & 2994          \\
    PseudoECA       & Cholec80      & \hspace{1ex}Algorithm\textsuperscript{a}         & 92,310        \\
    \end{tabular}
}
\tabnote{\textsuperscript{a}Handcrafted edge scoring variant of the algorithm presented in this paper}
\label{dataset-table}
\end{table}

\subsection{Estimation methods}

\subsubsection{Overview}
The task of estimating the content area comes down to correctly predicting the location and size of the circular image projection.
To reduce computational cost, we choose to investigate a number of discrete horizontal strips (i.e. small number of consecutive lines) from the image.
Two candidate edge points then need to be selected from each strip.
These candidate points can then be used to inform an estimation of the circular image projection.
Images with no clear circle must be assumed to be fully covered by the content area.
We present two different methods that follow this algorithm, which deviate only in how the edge scoring is performed.
The first method uses handcrafted features, whereas the second methods use a shallow CNN.


\subsubsection{Horizontal strip selection}
\begin{figure}[tb!]
\centering
\resizebox*{\textwidth}{!}{\includegraphics{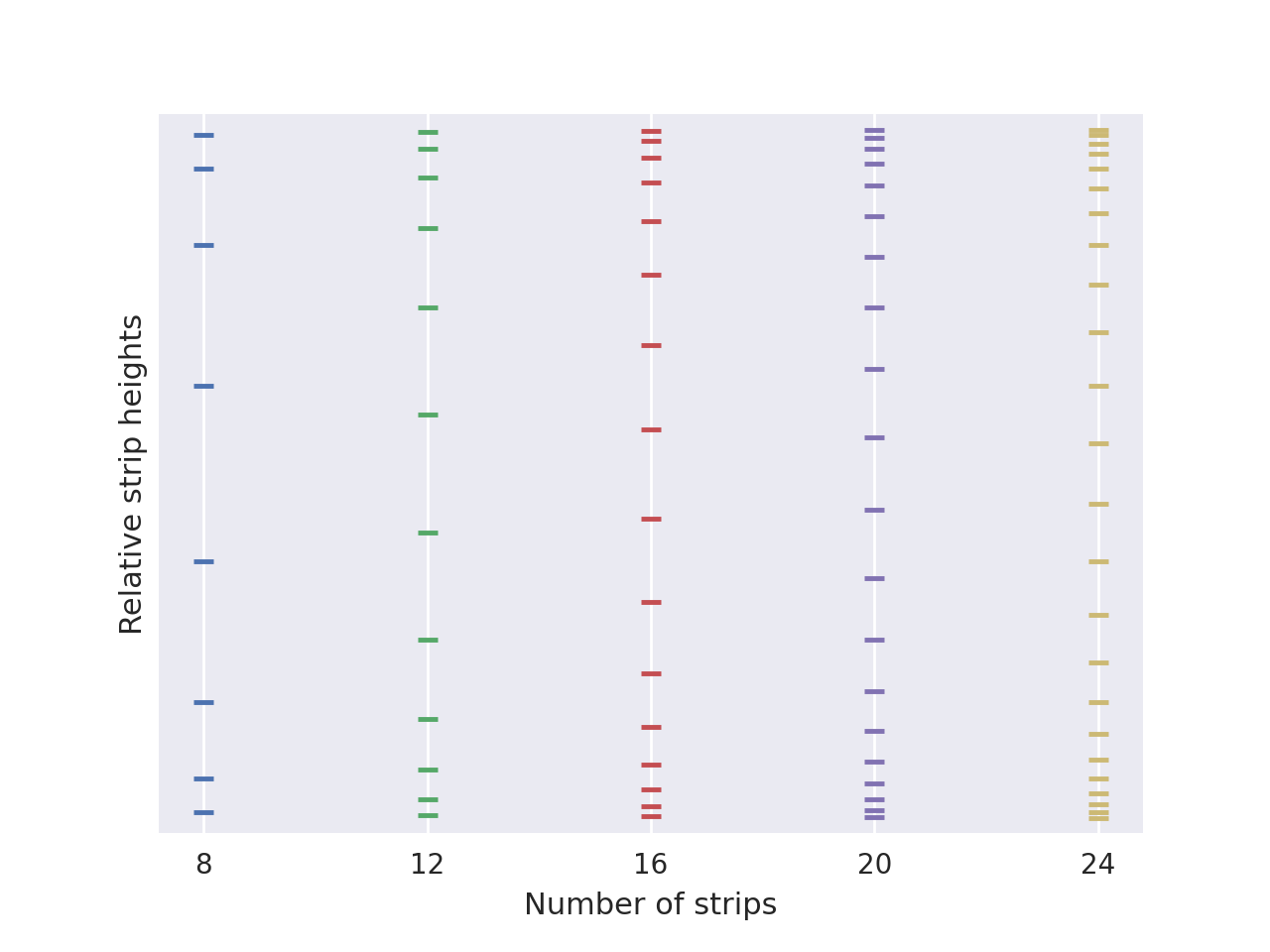}}
\caption{\revision[A demonstration of the relative strip heights calculated by equation~(\ref{eq:height}), with $\alpha=8$, for different numbers of strips.]} 
\label{fig:strip_heights}
\end{figure}

The obvious approach is to evenly space the horizontal strips across the vertical height of the image.
However, we find it productive to weight the vertical position of the strips to the top and bottom of the image.
The nature of the shape being detected makes the border much more prominent in these sections, and in many cases, the border is only visible at all in these sections.
For an image of height $H$, we choose the vertical position $h_i$ of strip $i$ of $N$ to be given by
\begin{equation}
    h_i = \frac{H}{1 + \exp\big({-\frac{\alpha}{N}(i - \frac{(N-1)}{2})}\big)}
    \label{eq:height}
\end{equation}
where $\alpha$ is a chosen value which affects how strongly the strips are weighted to the edges. 
We find $N=16$, and $\alpha=8$ to be a good configuration. \revision[Figure~\ref{fig:strip_heights} visually demonstrates the resulting strip heights for various number of strips.]

\subsubsection{Handcrafted edge scoring}
To craft a scoring metric, we must first select some image features to incorporate.
An obvious quality to work with is the gradient of the image $\vec{g}$ as calculated by applying 3$\times$3 Sobel filter.
The magnitude of the gradient $\|\vec{g}\|$ gives an obvious first feature.
We also note that, generally, the circular image projection is roughly centred in the image. This assumption allows us to approximate the vector $\vec{c}$ which points to the approximate centre circle.
Valid edges must have a gradient which points roughly in the same direction as $\vec{c}$.
It follows that the angle $\theta_g$ between $\vec{g}$ and $\vec{c}$ would make a good second feature.
Our final insight is that the intensity leading up to a valid edge should be quite dark.
To leverage this, we calculate the maximum intensity of the pixels preceding the currently investigated pixel in the strip, which we define as $\iota$.
For each of these three image features $\|\vec{g}\|$, $\theta_g$, and $\iota$, we manually define corresponding soft threshold values $t_g$, $t_\theta$, $t_\iota$.
We then form three scoring metrics which slowly saturate to $1$ or $0$ as the values move past these threshold values:
\begin{equation}
    s_{g} = \tanh\left(\frac{\|\vec{g}\|}{t_{g}}\right)
    ,\hspace{8pt}
    s_{\theta} = 1 - \tanh\left(\frac{\theta_g}{t_{\theta}}\right)
    ,\hspace{8pt}
    s_{\iota} = 1 - \tanh\left(\frac{\iota}{t_{\iota}}\right)
    \label{eq:induvidual_scores}
\end{equation}
The product of these then forms our final score:
\begin{equation}
    s = s_g s_\theta s_\iota
    \label{eq:final_score}
\end{equation}
We find $t_g=20$, $t_\theta=30^\circ$, and $t_\iota=25$ to be good values when considering pixels valued $0$ to $255$ and angles in degrees. \revision[A full parameter list is provided in the appendix.]

\subsubsection{Learned edge scoring}
We select a shallow CNN architecture, motivated by the simple nature of the problem and the desire to reduce computational cost. 
Our model consists of three 3$\times$3 convolutional layers with output features, $8$, $16$, and $32$.
Each convolution is followed by a ReLU activation layer. 
Finally, we perform a \revision[1$\times$1] convolution with a single output feature, which is then passed through a sigmoid layer \revision[to give a final probability estimates per pixel].

The input images are normalised using the mean and standard deviation of each colour channel calculated across the whole training dataset.
We also add the \revision[coordinate of each pixel] as two additional input features \revision[per pixel], which are also normalised \revision[to range from $-0.5$ to $0.5$ with $(0,0)$ corresponding to the centre of the image. The final image has 5 channels formatted as RGBXY.]

We use the pseudo-labelled PseudoECA dataset for training and the hand annotated data from CholecECA to validate.
In both cases, we generate edge maps in the form of crisp binary segmentation masks that only highlight the edge of the content area.
These edge maps are blurred using a Gaussian blur with a sigma of 3.
The motivation for this is twofold.
Firstly, learning to segment crisp fine structures is particularly challenging~\citep{xie15}.
Secondly, the pseudo-labelled, and even hand annotated, content areas may deviate slightly from the ground truth.

We choose to use the pseudo-labels to provide a substantially sized dataset, which would have taken a prohibitive amount of time to annotate. 
As the pseudo-labels are generated by the algorithm we hope to outperform, it remains to justify why the trained CNN would learn a more robust feature extraction.
We note that the extrapolation of the circle fitted to the found edge points discards edge points which may have been falsely detected, whilst also labelling edges which may have been missed or would have been missed if they fell on one of the inspected strips.

During training, the loss is calculated using soft binary cross entropy, and the model parameters are optimised using stochastic gradient descent with a learning rate of 0.001.
Images are fed to the network in batches of $8$.
We use the performance on our validation set to perform early stopping.

As part of the content area estimation algorithm, the trained CNN is applied to just the selected strips. 
The strips taken from the image must have a height of 7 pixels to match the patch size of the network. 
The resulting output probabilities are used as our final edge scores.

\subsubsection{Circle fitting}
We select a set of candidate edge points by taking the coordinates of the best scoring pixel from the left and right half of each strip.
We also note the score of each selected pixel to help inform the circle fitting.
We then use a combination of RANSAC \citep{fischler81} and least squares circle fitting \citep{coope93} to produce our final estimation of the circular image projection.

We start by filtering out points which lie within a manually defined number of pixels, $t_{px}$, from the image edge, or whose score falls below a second manually define value, $t_{ps}$.
Several triplets are selected, via uniform random sampling, which are used as the starting points for our RANSAC attempts.
Each attempt begins by analytically calculating the circle defined by the triplet.
We then define the inlier set as containing all the point which fall within a defined number of pixels, $t_{ri}$, from the triplet-defined circle.
Least squares fitting is then used to fit a circle to the current inlier set.
The circle is scored by summing the edge scores of the inlier set.
The inlier set can then be updated, and the process can be repeated, gradually refining the fit, although we find $3$ iterations to be sufficient.

The final circle candidates from each RANSAC attempt are filtered to remove those whose radius falls outside the bounds $r_{min}$ and $r_{max}$, and for whom the separation from the circle centre to the image centre is greater than $d_{max}$.
The scores from each RANSAC attempt are compared, and the best scoring circle is taken as the final estimation of the circular image projection.
If the score of the final circle falls below a manually defined threshold, $t_{cs}$, The circle is rejected and the content area is assumed to take up the whole image rectangle. 

We find the following thresholds to work well: $t_{px}=3$, $t_{ps}=0.03$, $t_{ri}=3$, and $t_{cs}=0.06$. As for the limits on the circle parameters, we choose permissive values expressed as factors of the image width: $r_{min}=0.1$, $r_{max}=0.8$, and $d_{max}=0.2$. \revision[A full parameter list is provided in the appendix.]

\subsubsection{GPU parallelisation summary}
To reduce runtime, the entire algorithm is implemented on the GPU and in a way that tries to maximise parallelism.
Here we provide a summary of the parallelisation scheme adopted to achieve this.

For the handcrafted edge scoring, strips are divided into the left-hand and right-hand side.
Each half-strip is processed in parallel by separate groups of threads known as ``blocks'' which are \revision[capable of] sharing information between threads during processing.
Each block is allocated enough threads for each thread to process an individual pixel.
The two non-trivially parallelised operations are the calculation of the preceding maximum intensity and the search for the maximum scoring pixel.
These two operations are achieved using a block-wise parallel scan and reduction respectively, operations which will be familiar to those with GPU programming experience. 

Finally, the circle fitting is performed by a single block of threads, with each individual thread handle a single RANSAC attempt. 
This allows for many attempts to be made in parallel, thus efficiently increasing the chances of finding an optimal fit.
Similarly to finding the best scoring edge points, finding the best scoring circle is handled by a block-wise parallel reduction. 

\subsection{Evaluation methodology}
For evaluation of content area estimation methods, we propose using the Hausdorff distance \citep{huttenlocher93}. Given two sets of points $A=\{a_1,a_2,...\}$, and $B=\{b_1,b_2,...\}$, the Hausdorff distance is defined as
\begin{align}
    H(A, B) &= \max(h(A, B), h(B, A)) \quad \textrm{such that} \\
    \label{eq:hausdorf}
    h(A, B) &= \max_{a \in A}\min_{b \in B} f(a,b)
\end{align}
where $f(a,b)$ is some distance function between two points. 
In essence, it calculates the furthest distance from any point in either set to its closest neighbour in the other set. For our use case, we define the two sets as the points along the edges of the ground truth and inferred content areas, and the two-dimensional Euclidean distance is used to calculate the distances between points.
Furthermore, in order to normalise scores between different images, we choose to scale the distances by the ratio of the diagonal length $d_I$ of the image with the diagonal length $d$ of a 1080$\times$1920 image. The resulting normalised Hausdorff distance is defined as
\begin{equation}
    NH(A, B) = \frac{d}{d_I} H(A,B)
    \label{eq:nhausdorf}
\end{equation}
We also define two thresholds for the Normalised Hausdorff distance, beyond which the predicted content area is considered to be a miss, and a bad miss. We find 15 px and 25 px to be good cut off points.






\section{Results}
\begin{figure}[tb!]
\centering
\subfloat[The dark right-hand side of the content area is missed by both variants. The learned variant gets a slightly better fit, this could be due to a better positioing of points on the left-hand side of the content area.]{%
\resizebox*{0.48\textwidth}{!}{\includegraphics{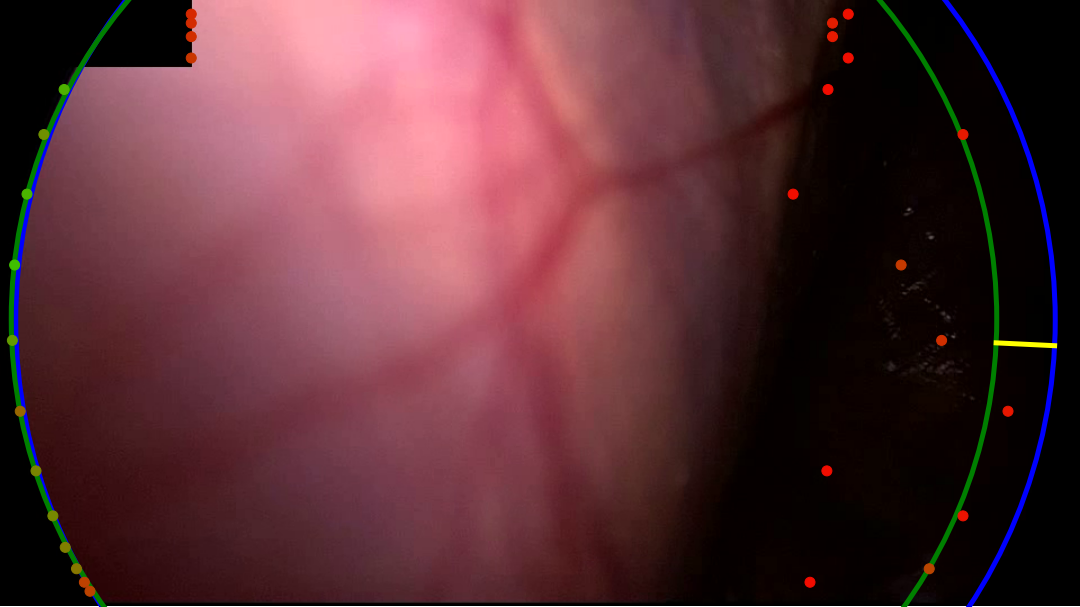}}
\resizebox*{0.48\textwidth}{!}{\includegraphics{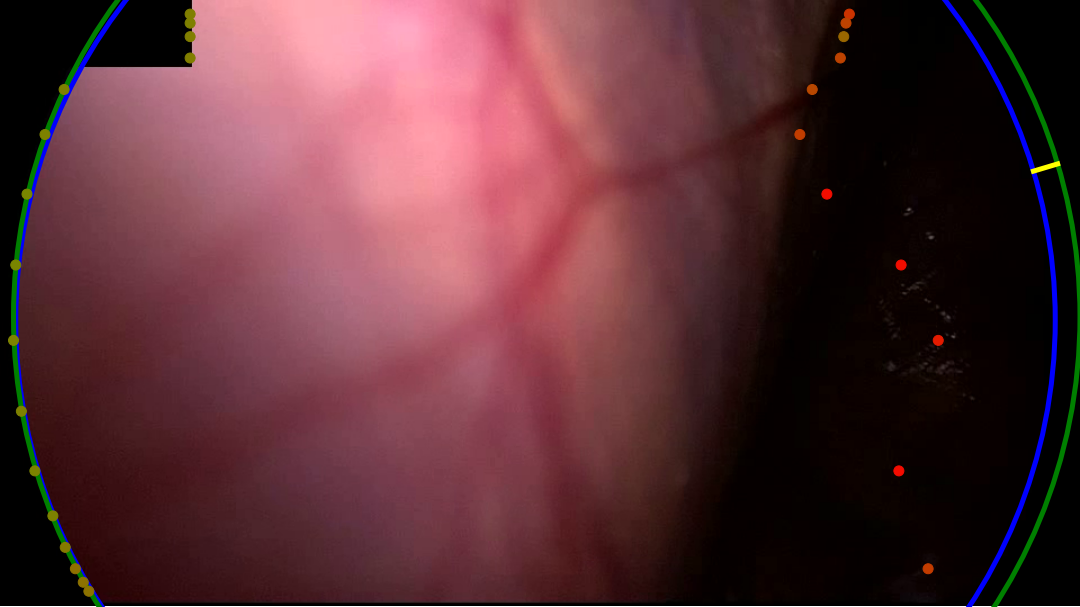}}} \\
\vspace{-1ex}
%
\subfloat[A noisy and bright border poses a challenge for the handcrafted variant. Too few edge points are detected to make a confident estimate of the circle. The learned variant manages to correctly identify the right-hand edge of the content area, however, the circle fits to incorrect edge points found on the blacked out secondary video feed in the top-left.]{%
\resizebox*{0.48\textwidth}{!}{\includegraphics{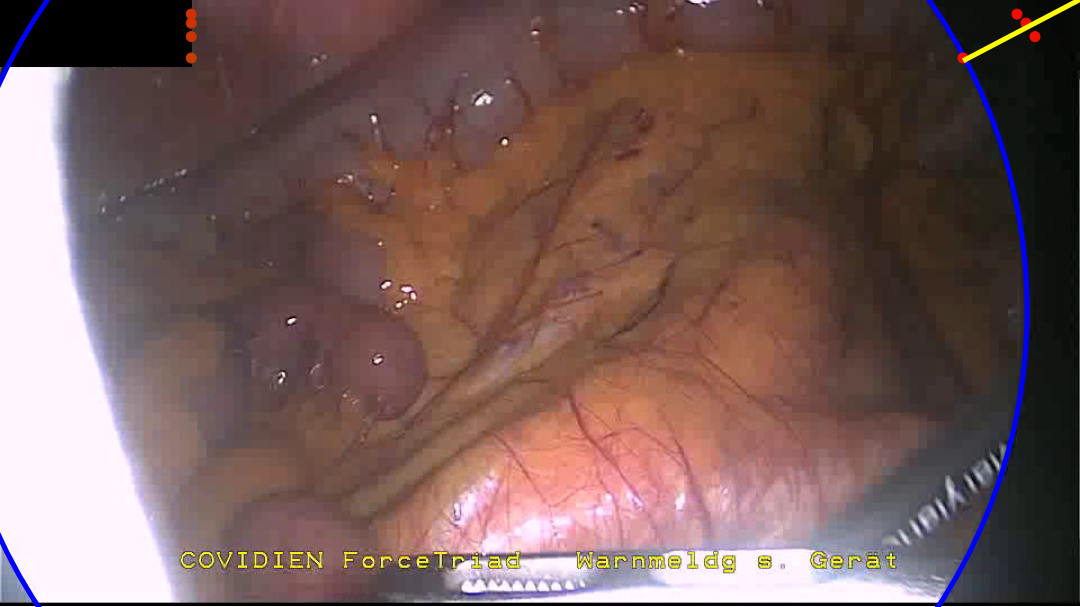}}
\resizebox*{0.48\textwidth}{!}{\includegraphics{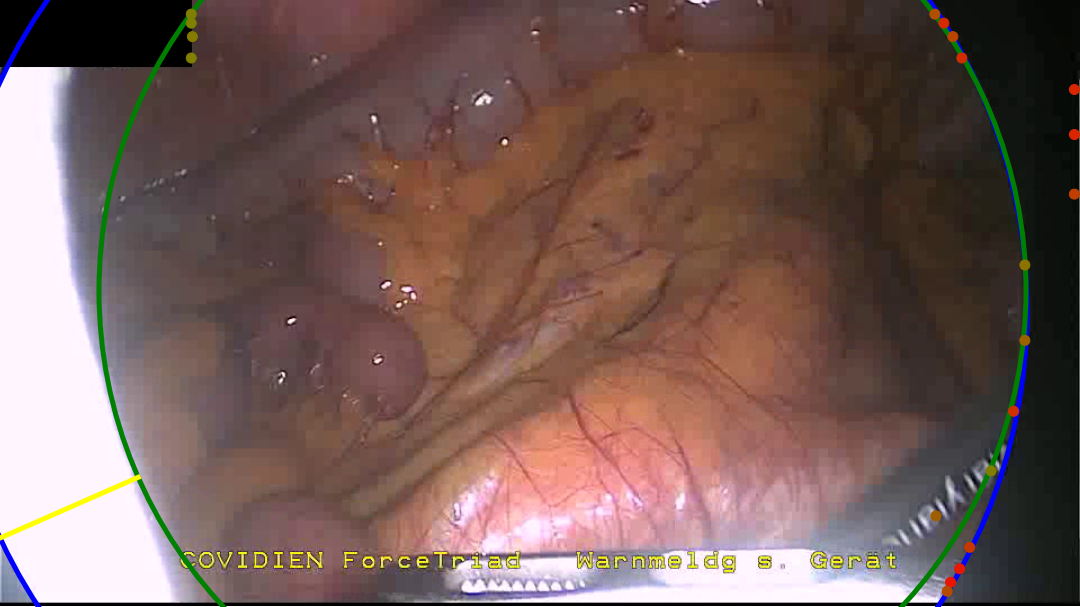}}} \\
\vspace{-1ex}
%
%
\subfloat[Both variants correctly identify the edge in the bottom-left but mistake dark regions in the top-right and bottom-right corners as border regions. The learned variant rejects the final circle as the circle score is too low. This scenario, where a single corner of the image is outside the content area, features prominantly in our failure cases.]{%
\resizebox*{0.48\textwidth}{!}{\includegraphics{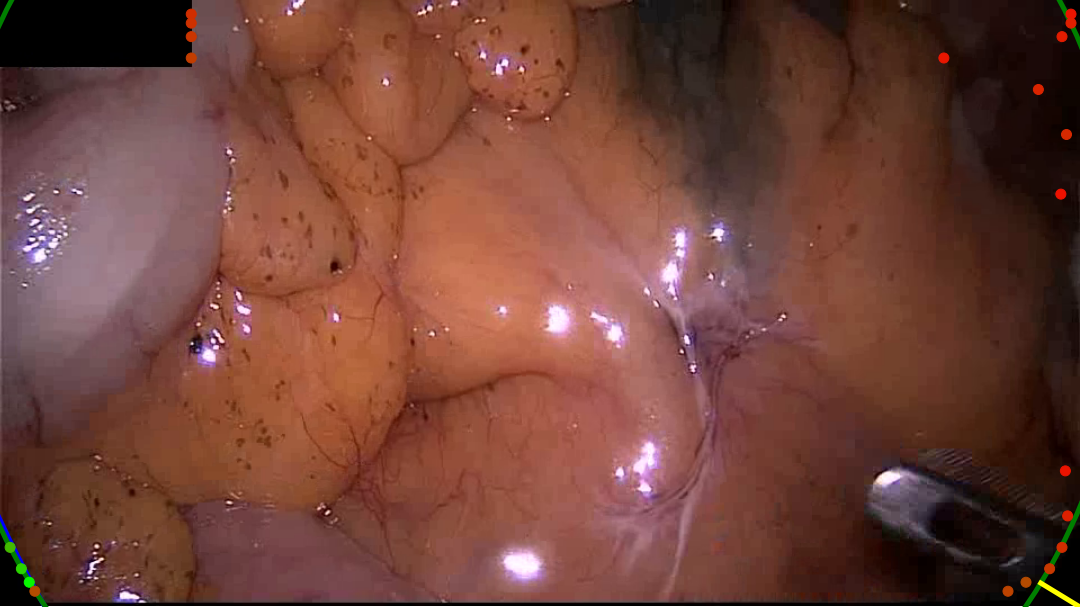}}
\resizebox*{0.48\textwidth}{!}{\includegraphics{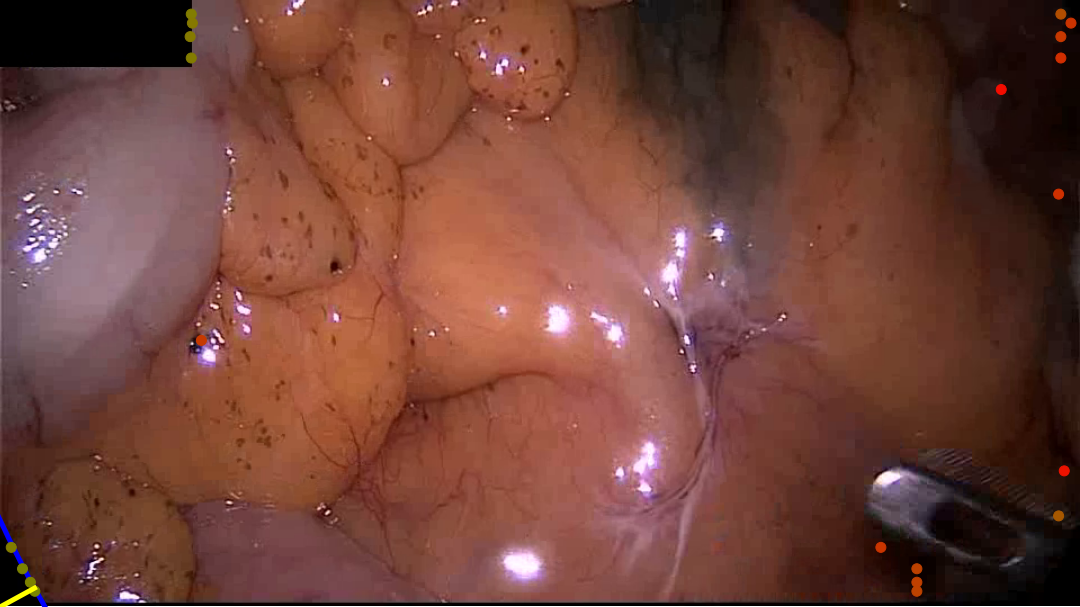}}} \\
\vspace{-1ex}
\subfloat[In this sample, the handcrafted variant detects edge points in the bottom-left, but scores them poorly. The circle fitting then chooses the incorrectly identified edge points in the top-left. The learned variant scores the detected points in the bottom-left higher, and so the circle fitting finds a satisfying result.]{%
\resizebox*{0.48\textwidth}{!}{\includegraphics{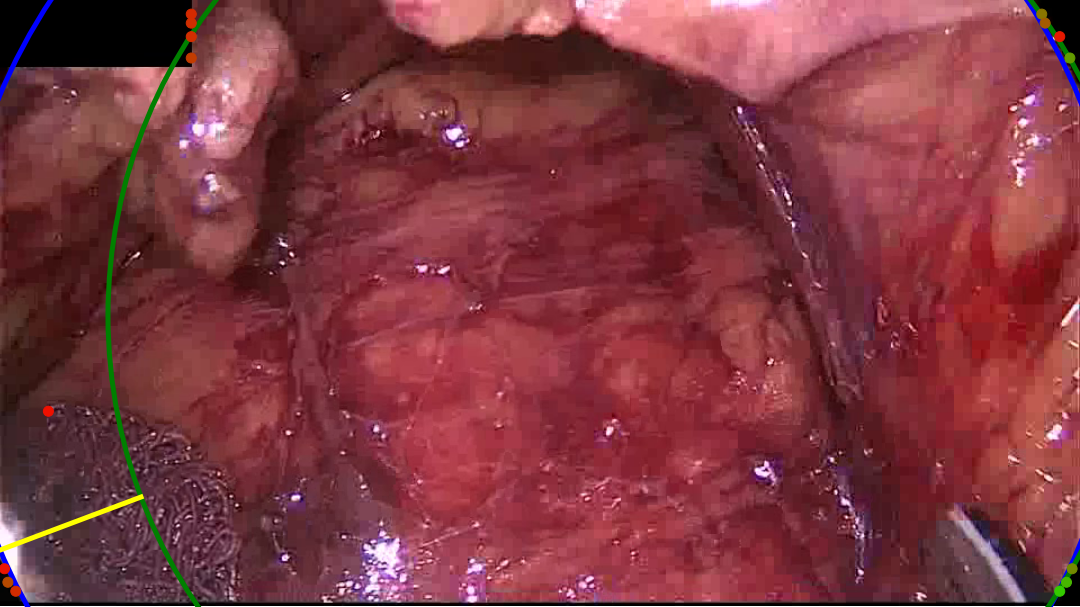}}
\resizebox*{0.48\textwidth}{!}{\includegraphics{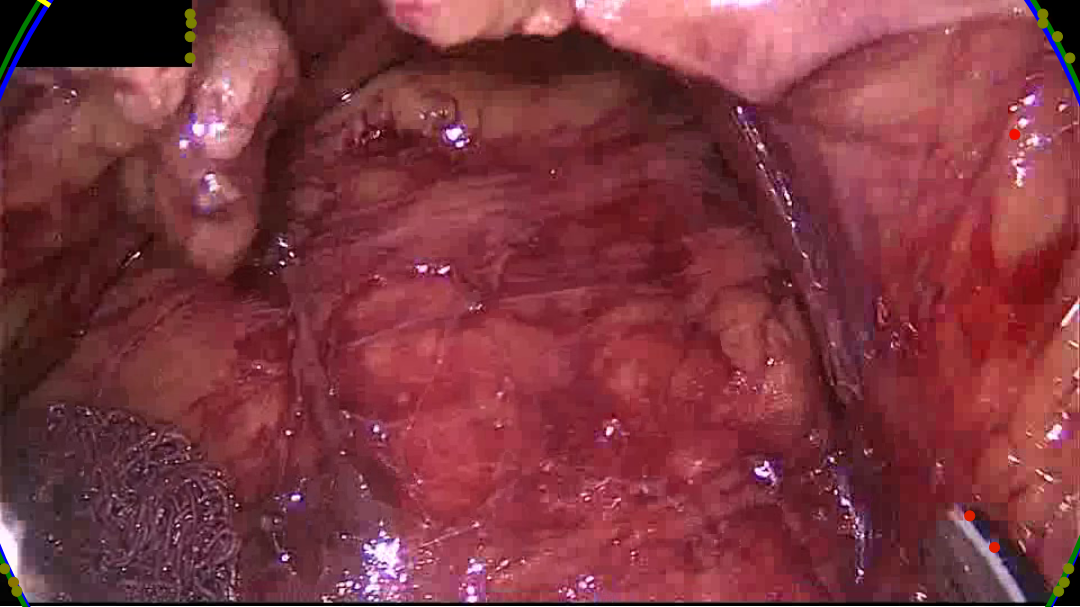}}} \\
\vspace{-1ex}
\caption{A selection of failure cases of our algorithm taken from the RobustECA dataset, the results of both the handcrafted (left) and learned (right) variants are shown for each image. The edge of the ground truth and inferred circular image projections are shown in blue, and green, respectively. Points indicate the position of detected edge points, the colour of each point indicates its score, going from red to green as the score increases. The vector found when calculating the Hausdorff distance is shown in yellow.} 
\label{fig:failures}
\end{figure}

Table~\ref{results-table} shows the scores achieved both the handcrafted and learned variants of our algorithm. 
Also shown are the results for two publicly available content area estimation algorithms, one available in the PyPI package ``endoscopy'' \citep{huber22}, and a second available in the PyPI package ``endoseg'' \citep{gruijthuijsen2022}. 
Additionally, we re-implemented the algorithm presented by \cite{munzer13}. The description of the algorithm lacks some critical details, particularly various parameter values.
As our two algorithm variants were developed and tuned against the CholecECA dataset, the scores achieved on the CholecECA and RobustECA datasets are provided separately. 
In this way, the two datasets play the roles of validation and testing sets. Methods which we did not develop are also presented in this way for comparison purposes.
Figure~\ref{fig:failures} shows a selection of the failure cases, showing the results for both variants of our algorithm.

\begin{table}[htb!]
\tbl{The results achieved for each method run on both the CholecECA, and RobustECA Datasets. The runtimes for each method, acheived by averaging over a thousand runs on an NVIDIA GeForce GTX 980 Ti, are also shown.}
{

    \begin{tabular}{lcccccccc} 
    \toprule
    \multirow{3}{*}{Method} & \multicolumn{3}{l}{CholecECA} && \multicolumn{3}{l}{RobustECA} & \multirow{3}{*}{Runtime (ms)}\\
    \cmidrule{2-4}\cmidrule{6-8}
                        & Avg. err. & Miss & Bad Miss && Avg. err. & Miss & Bad Miss\\
                        & (px) & (\%) & (\%) && (px) & (\%) & (\%)\\ 
    \midrule
    
    \cite{munzer13}\textsuperscript{a} & 69.19 & 22.5 & 22.0 && 56.70 & 19.3 & 19.3 & \hspace{0.5ex}6.211\textsuperscript{d}\\
    
    Endoseg\textsuperscript{b} & 63.93 & 21.9 & 18.3 && 34.67 & 22.7 & 12.2 & 40.065\\
    
    Endoscopy\textsuperscript{c} & 96.47 & 12.7 & 11.0 && 118.14 & 16.3 & 14.4 & 11.198\\
    
    Handcrafted (ours) & \bftab{3.70} & \bftab{2.0} & \bftab{1.1} && \bftab{6.30} & 2.8 & 2.5 & \bftab{0.129}\\
    
    Learned (ours) & 4.36 & 2.4 & 1.2 && 6.48 & \bftab{2.7} & \bftab{2.3} & 2.542\\
    
    \bottomrule
    \end{tabular}
}
\tabnote{\textsuperscript{a}our re-implementation following the work by \cite{munzer13}}
\tabnote{\textsuperscript{b}\url{https://github.com/luiscarlosgph/endoseg} \citep{gruijthuijsen2022}}
\tabnote{\textsuperscript{c}\url{https://github.com/RViMLab/endoscopy} \citep{huber22}}
\tabnote{\textsuperscript{d}runtime taken from original work \citep{munzer13}}
\label{results-table}
\end{table}

\section{Discussion}

The dataset produced during this work covers a wide range of difficult cases which should be handled during content area detection, and should serve as a solid baseline.
We do note however that the data is taken solely from laparoscopic video.
While it is arguable that, assuming a circular content area, the task would not change significantly between types of endoscopy, it would still be better to incorporate other sources of endoscopic video, particularly types of flexible endoscopy such as colonoscopy, gastroscopy, or vascular endoscopy.

From the results presented in Table~\ref{results-table}, we see that our two approaches significantly outperform existing available techniques, both in accuracy, robustness, and runtime.
Indeed, our handcrafted variant runs two orders of magnitude faster than the fastest existing method by \cite{huber22} while improving the percentage of misses by at least a factor 4.

Comparing our two algorithm variants, we see that the handcrafted approach performs significantly better on the CholecECA validation dataset, but the performance gap is not significant on the RobustECA testing dataset, with the learned variant even beating the handcrafted variant when considering the rate of misses and bad misses.
This could suggest that the threshold values in the handcrafted approach are easily over tuned, whereas the CNN has managed to learn a more generalisable feature extraction.
It is also worth noting that the handcrafted variant is provided the preceding maximum pixel intensity, information which would not be available to the CNN.
These two factors, along with a lack of experimentation with model architectures and training methodologies, suggest that a learned feature extraction could perform better with a more thorough investigation.

Inspecting some of the results, a few of which are shown in Figure~\ref{fig:failures}, we see that we fail most often when the content area makes up the whole image except for a few corners which lie just out-side of the circular image projection. One issue in particular arises when this is combined with the black box in the top-left, which is present in a number of the samples from the RobustECA dataset.

An aspect which we did not consider is the use of temporal information.
Real-time content area estimation could benefit from the use of the previous frames, whereas pre-recorded video could be processed using information from both previous and subsequent frames. 
A simplistic solution would be a temporal filtering of the estimated content area, perhaps taking into account the circle score as a confidence measure. 
A more in-depth solution could be to use the previously estimated content area to directly inform the edge point scoring, and circle fitting process.


An implementation of our algorithm with a PyTorch binding has been published as a PyPI package, the source code for which has been made available \anoncmd{under an MIT licence on GitHub\footnote{\url{https://github.com/charliebudd/torch-content-area}}}{at \url{anonymized url}}.
In addition, the curated datasets, together referred to as the endoscopic content area~(ECA) dataset, has been published \anoncmd{to synapse\footnote{\url{https://doi.org/10.7303/syn32148000}}}{at \url{anonymized url}} to encourage further developments.




\FloatBarrier

\anoncmd{
\section*{Acknowledgements}
This work was supported by core and project funding from the Wellcome/EPSRC [WT203148/Z/16/Z; NS/A000049/1; WT101957; NS/A000027/1].
This project has received funding from the European Union's Horizon 2020 research and innovation programme under grant agreement No 101016985 (FAROS project).
TV is supported by a Medtronic / RAEng Research Chair [RCSRF1819\textbackslash7\textbackslash34].
For the purpose of open access, the authors have applied a CC BY public copyright licence to any Author Accepted Manuscript version arising from this submission.

\section*{Disclosure statement}
TV and SO are co-founders and shareholders of Hypervision Surgical.
TV also holds shares from Mauna Kea Technologies.
}{}

\bibliographystyle{tfcse}
\bibliography{library}

\newpage

\appendix

\section{Parameters}

\begin{table}[htb!]
\tbl{\revision[A tabulation of the parameters for our content area estimation methods. The Parameters are presented in two blocks. The upper block are the main parameters of the methods and may be of interest to end users whishing to tune, although the default values should be sufficient in most use cases. The second block of parameters are provided for completeness but are likely of little interest to end users.]}
{
    \begin{tabular}{cccp{0.6\linewidth}} 
    \toprule
    Parameter & Value & Method & Description\\
    \midrule
    $N$ & 16 & both & The number of image strips to investigate.\\
    $t_g$ & 20 & handcrafted & Gradient magnitude threshold for edge point scoring.\\
    $t_\theta$ & 30 & handcrafted & Gradient direction threshold for edge point scoring.\\
    $t_\iota$ & 25 & handcrafted & Preceding maximum intensity threshold for edge point scoring.\\
    $t_{ps}$ & 0.03 & both & Point score threshold to filter low scoring points.\\
    $t_{cs}$ & 0.06 & both & Final circle score threshold to filter out low scoring circles.\\
    \midrule
    $\alpha$ & 8 & both & Determines to what extent strips are weighted to the vertical extremes of the image.\\
    $t_{px}$ & 3 & handcrafted & Candidate edge points within this many pixels of the border are discarded.\\
    $t_{ri}$ & 3 & both & The distance in pixels within which a candidate edge point is counted as an inlier for the RANSAC circle fitting\\
    $r_{min}$ & 0.1 & both & Minimum radius of a found circle (multiple of image width).\\
    $r_{max}$ & 0.8 & both & Maximum radius of a found circle (multiple of image width).\\
    $d_{max}$ & 0.2 & both & Maximum distance between the center of the image and a found circle (multiple of image width).\\
    \bottomrule
    \end{tabular}
}
\end{table}

\end{document}